\pdfoutput=1
\documentclass[11pt]{article}

\usepackage[]{acl}

\usepackage{times}
\usepackage{latexsym}

\usepackage[T1]{fontenc}
\usepackage[utf8]{inputenc}
\usepackage{caption}
\usepackage{subcaption}
\usepackage[colorinlistoftodos]{todonotes}
\usepackage{makecell}
\usepackage[ruled,vlined,linesnumbered]{algorithm2e}
\usepackage{amsfonts}
\usepackage{amsmath}

\usepackage{microtype}

\title{Frame Shift Prediction}

\author{Zheng-Xin Yong\thanks{Work done when studying at Minerva University in 2020.} \and Patrick D. Watson \\ Minerva University \\ \texttt{\{zhengxin.yong, pwatson\}}\\\texttt{@uni.minerva.edu}
        \And
        Tiago Timponi Torrent \\ Federal University of Juiz de Fora \\ \texttt{tiago.torrent@ufjf.edu.br}
        \AND
        Oliver Czulo \\ University of Leipzig \\ \texttt{czulo@uni-leipzig.de}
        \And
        Collin F. Baker \\ International Computer Science Institute \\ \texttt{collinb@icsi.berkeley.edu}
        }

\begin{document}
\maketitle
\begin{abstract}
\textit{Frame shift} is a cross-linguistic phenomenon in translation which results in corresponding pairs of linguistic material evoking different frames. The ability to predict frame shifts enables automatic creation of multilingual FrameNets through annotation projection. Here, we propose the Frame Shift Prediction task and demonstrate that graph attention networks, combined with auxiliary training, can learn cross-linguistic frame-to-frame correspondence and predict frame shifts.
\end{abstract} 

\section{Introduction}
Frame Semantics \citep{fillmore82frame} is an approach to meaning that characterizes the background knowledge against which linguistic entities are understood via systems of interrelated concepts called \textit{frames}. Frames are evoked by \textit{lexical units} (LUs) and involve participants and props characterized as \textit{frame elements}. The Berkeley FrameNet (BFN) project \citep{ruppenhofer2016framenet} establishes a general-purpose resource for frame semantic descriptions of English and is widely adopted for many NLP applications. BFN has successfully been adapted for other languages such as Brazilian Portuguese \citep{torrent2013behind,torrent2018fnbr} and German \citep{burchardt-etal-2006-salsa,boas2018constructing}. The success of these works implies that some frames are applicable across different languages, while others can be adapted to fit language specificities \citep{gilardi2018multilingual,baker-lorenzi-2020-exploring}.

\begin{figure}[htp]
    \centering
    \includegraphics[width=6cm]{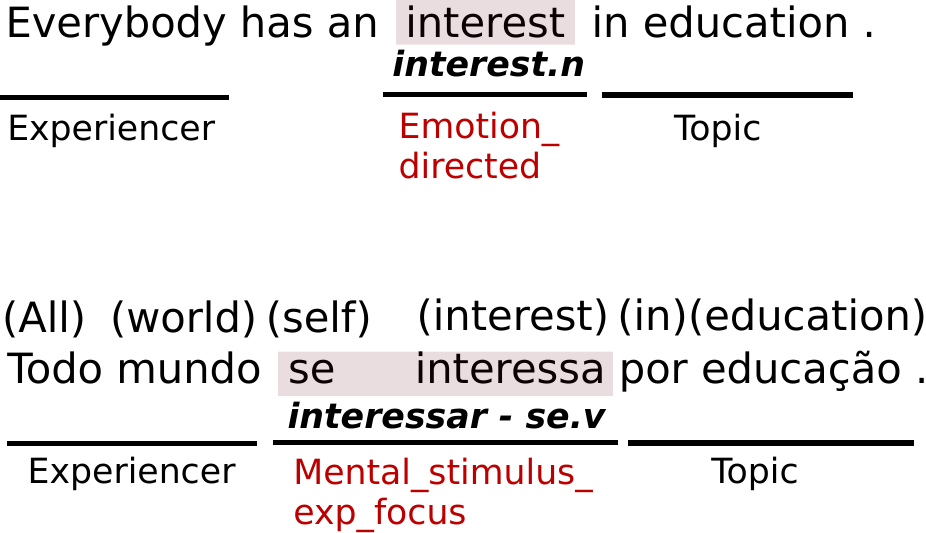}
    \caption{An example of frame shift in a pair of English and Brazilian Portuguese parallel sentences. The frame-evoking lexical units are bold and italicized, and the frames are colored in red.}
    \label{fig:frame_shift}
\end{figure}

Nonetheless, researchers working with non-English FrameNets find that differences in lexical-constructional patterns between languages result in \textit{frame shifts} \cite{subirats2003surprise,litkowski2009hans,pado2009cross,czulo2017aspects,linden2019finntransframe,giouli-etal-2020-greek,ohara-2020-finding}. Because FrameNet frames are sensitive to valency variations, when a given sentence is translated from one language to another, it is usually the case that the frames evoked will be significantly different. \citet{torrent2018multilingual} show that the average direct frame correspondence between sentences in English and their translation to Brazilian Portuguese is only 0.51. Here, as shown in Figure~\ref{fig:frame_shift}, we focus on a type of frame shift where, in a pair of parallel sentences, the corresponding LUs evoke two different frames. The issue of frame shift is of major importance for projecting annotations from one language to another and for bootstrapping FrameNets from parallel corpora, and we address this issue by using graph neural networks to predict frame shifts.

In summary, this paper makes the following contributions: (a) we present a case study of frame shifts and identify their possible causes; (b) we propose a new task, Frame Shift Prediction (FSP), for predicting frame shifts in parallel texts; (c) our empirical results show that representing frames with graph attention networks (GAT) outperforms existing multilingual frame embedding methods in FSP. To the best of our knowledge, we are the first to use GAT to represent frames.

\section{Related Work}
\paragraph{Multilingual FrameNet.} Currently, there are more than 20 FrameNet (FN) resources in various languages \cite{gilardi2018multilingual,giouli-etal-2020-greek,gargett-leung-2020-building}. One way to automatically expand FN to a new language is by projecting frame annotations \cite{johansson-nugues-2006-framenet}. While FNs in each language use similar or even identical frames for annotation, recent work reports variations in how the translators "frame" the sentences on the conceptual level \cite{subirats2003surprise,litkowski2009hans,pado2009cross,vculo2013constructions,linden2019finntransframe,giouli-etal-2020-greek,ohara-2020-finding}.

\paragraph{Graph Neural Networks.} Graph neural networks (GNNs) have been proven successful in encoding relational data in NLP tasks, such as knowledge graph completion \cite{shang2019end,nathani-etal-2019-learning}, syntactic and semantic parsing \cite{marcheggiani-titov-2017-encoding,bogin-etal-2019-representing,ji-etal-2019-graph}. Modeling FN-specific information is no exception:  \citet{li2017situation} and \citet{suhail2019mixture} applied GNNs to learn the dependencies between verb and frame-semantic roles for situation recognition. Here, we use the graph attention network \cite{velivckovic2018graph} to model frame relations and predict frame shifts.

\section{Linguistic and Quantitative Analysis of Frame Shifts}
\label{sec-shift-analysis}
\subsection{Frame Shifts Dataset}
\label{sec:frame-shifts-dataset}
We create the dataset for frame shifts from the Global FrameNet Shared Annotation Task \cite{torrent2018multilingual}, which has been devised to assess whether frames in BFN 1.7 suit the semantics of LUs in different languages. Given an English sentence and its translation in German (DE) and Brazilian Portuguese (PT), we first construct the word-to-word correspondence with Fast Align \citep{dyer-etal-2013-simple} and then extract the corresponding LUs. We also use the Open Multilingual WordNet \cite{Bond:Paik:2012} to filter out false positives, that is, LUs which are not translation equivalents. In the end, we extract 95 EN-DE and 316 EN-PT annotation pairs for FSP. Frame shifts are found in 36\% of the EN-DE and 22.4\% of the EN-PT pairs.

\subsection{Translational divergences}
\label{sec:translational-divergences}
Translational divergences are described in \citet{dorr-1994-machine}. Next, we show how each of them leads to frame shifts in our dataset. 

\paragraph{Categorial} 
Categorial divergence happens when two languages use words of different parts-of-speech to express the same meaning. Figure \ref{fig:frame_shift} illustrates the example of frame shift caused by categorial divergence: the English noun \textit{interest} in the phrase \textit{has an interest} corresponds to the Portuguese verb phrase \textit{se interessa ("to interest oneself")}. They evoke different frames as the former refers to the feeling of interest, whereas the latter refers to the evocation of an emotional response in the \textsc{Experiencer} to the \textsc{Topic}.

% \begin{figure}[htp]
%     \centering
%     \includegraphics[width=6cm]{assets/categorical_divergence.pdf}
%     \caption{Frame shift due to categorial divergence.}
%     \label{fig:categorial_dvg}
% \end{figure}

\begin{figure}[htp]
    \centering
    \includegraphics[width=4.5cm]{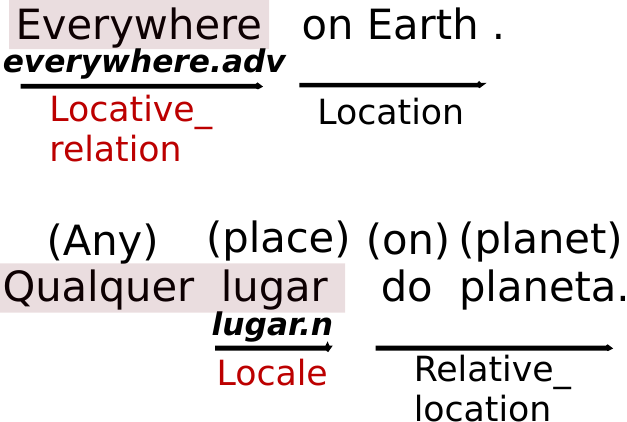}
    \caption{Frame shift due to inflational divergence.}
    \label{fig:inflational_dvg}
\end{figure}

\paragraph{Conflational/Inflational} Conflational divergence occurs when two or more words in one language are translated into one word in another language, whereas inflational divergence is the opposite. In Figure \ref{fig:inflational_dvg}, the inflational divergence splits the English word \textit{everywhere} into two Portuguese words and causes the translated frame-evoking counterpart \textit{lugar (place)} to lose the conceptual relativity to other locations.

\begin{figure}[htp]
    \centering
    \includegraphics[width=5cm]{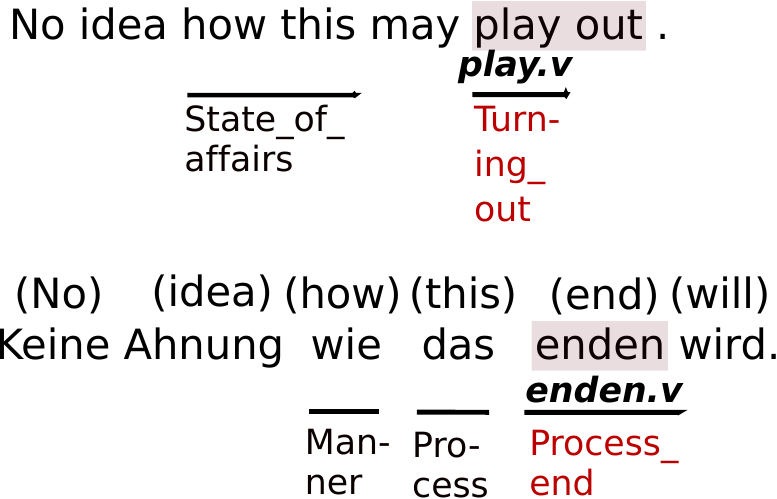}
    \caption{Frame shift due to lexical divergence.}
    \label{fig:lexical_dvg}
\end{figure}

\paragraph{Lexical} The unavailability of an exact translation for a construction in a language leads to lexical divergence. In frame semantics, divergence of LUs often causes divergence in meaning. For instance, as shown in Figure \ref{fig:lexical_dvg}, the German translation for the English phrase \textit{play out} is \textit{enden (end)}, which differs on the dimension of the realization of the terminal state.

\begin{figure}[htp]
    \centering
    \includegraphics[width=6cm]{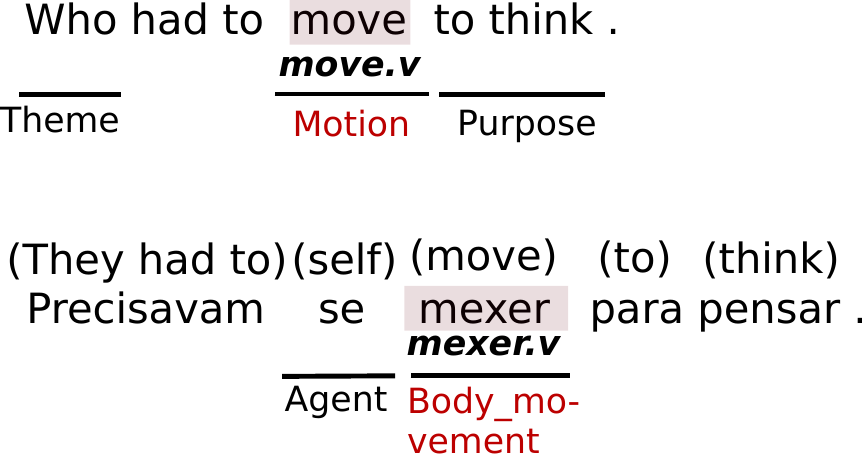}
    \caption{Frame shift due to structural divergence.}
    \label{fig:structural_dvg}
\end{figure}

\paragraph{Structural} This divergence happens when verb arguments result in different syntactic configurations. In Figure~\ref{fig:structural_dvg}, the English verb \textit{move} does not take a reflexive direct object so it evokes the \texttt{Motion} frame, in which the subject of motion is the \textsc{Theme} (entity that changes location). On the other hand, the Portuguese verb \textit{mexer (move)} is adjacent to the reflexive particle \textit{se (self)}, which is interpreted as the \textsc{Agent} moving his/her body; therefore, the verb evokes the \texttt{Body\_movement} frame.

\paragraph{Thematic and Head Swapping}  We did not observe frame shifts caused by thematic divergence (inversion of semantic roles) and head swapping (inverted direction of the dependency relations) in our dataset. 

\subsection{Construal Differences}
We find that translational divergences alone (Section~\ref{sec:translational-divergences}) are insufficient to account for all instances of frame shift. The reason is that the linguistic expressions can be almost identical in the semantic content but differ in the decoding of their meanings along certain dimensions of construal \cite{verhagen2007construal, trott-etal-2020-construing}. The following analysis of the effects of construal operations is by no means exhaustive.

\begin{figure}[htp]
    \centering
    \includegraphics[width=7.5cm]{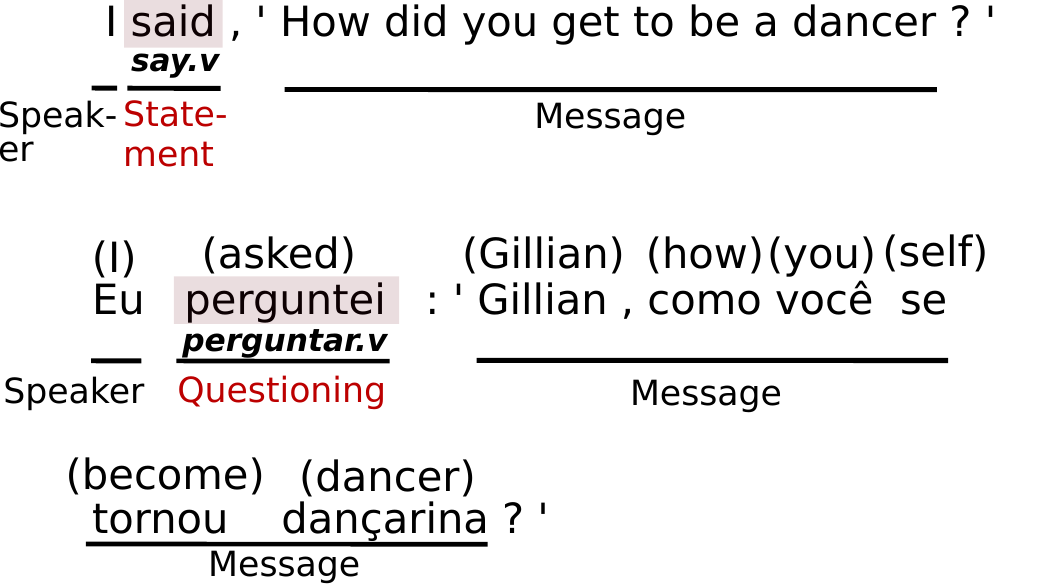}
    \caption{Frame shift due to differences in resolution.}
    \label{fig:resolution_dvg}
\end{figure}

\paragraph{Resolution} As lexical categories form taxonomic hierarchies consisting of various levels of specificity (e.g., cat $<$ mammal $<$ animal $<$ organism), language users can express a concept with different degrees of granularity. Therefore, the expressions can evoke different frames. In Figure \ref{fig:resolution_dvg}, the lexical item \textit{said} is more schematic compared to the word \textit{perguntei (ask)} where the former denotes the generic action of communicating a message, whereas the latter provides additional information about the nature of the message. 

%\paragraph{Perspective} Languages have different ways to introduce relationships among entities. In Frame Semantics, perspective captures the different points of view which may be adopted while referring to one same scene. In Figure X, the English sentence evokes the texttt{Mental\_stimulus\_experiencer\_focus} frame, while, in the German translation, the perspective taken is that of the \textsc{Stimulus}.

\begin{figure}[htp]
    \centering
    \includegraphics[width=4cm]{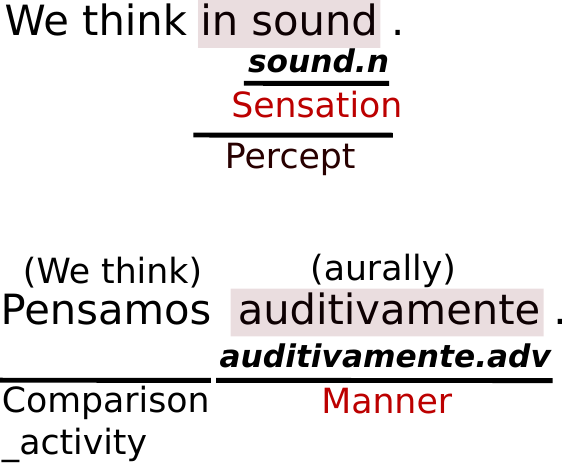}
    \caption{Frame shift due to differences in prominence.}
    \label{fig:prominence_dvg}
\end{figure}

\paragraph{Prominence} Prominence refers to the relative focus of attention on elements against the rest in a scene. In Figure \ref{fig:prominence_dvg}, while both sentences characterize the style of thinking with adverbial phrases, the linguistic expression \textit{in sound} makes explicit the auditory sensation. In contrast, the Portuguese adverb \textit{auditivamente (aurally)} foregrounds the thinking action \textit{Pensamos (We think)}, which is labeled with the frame element \textsc{Comparison\_activity} that indicates the activity characterized by the \texttt{Manner} frame.

\begin{figure}[htp]
    \centering
    \includegraphics[width=8cm]{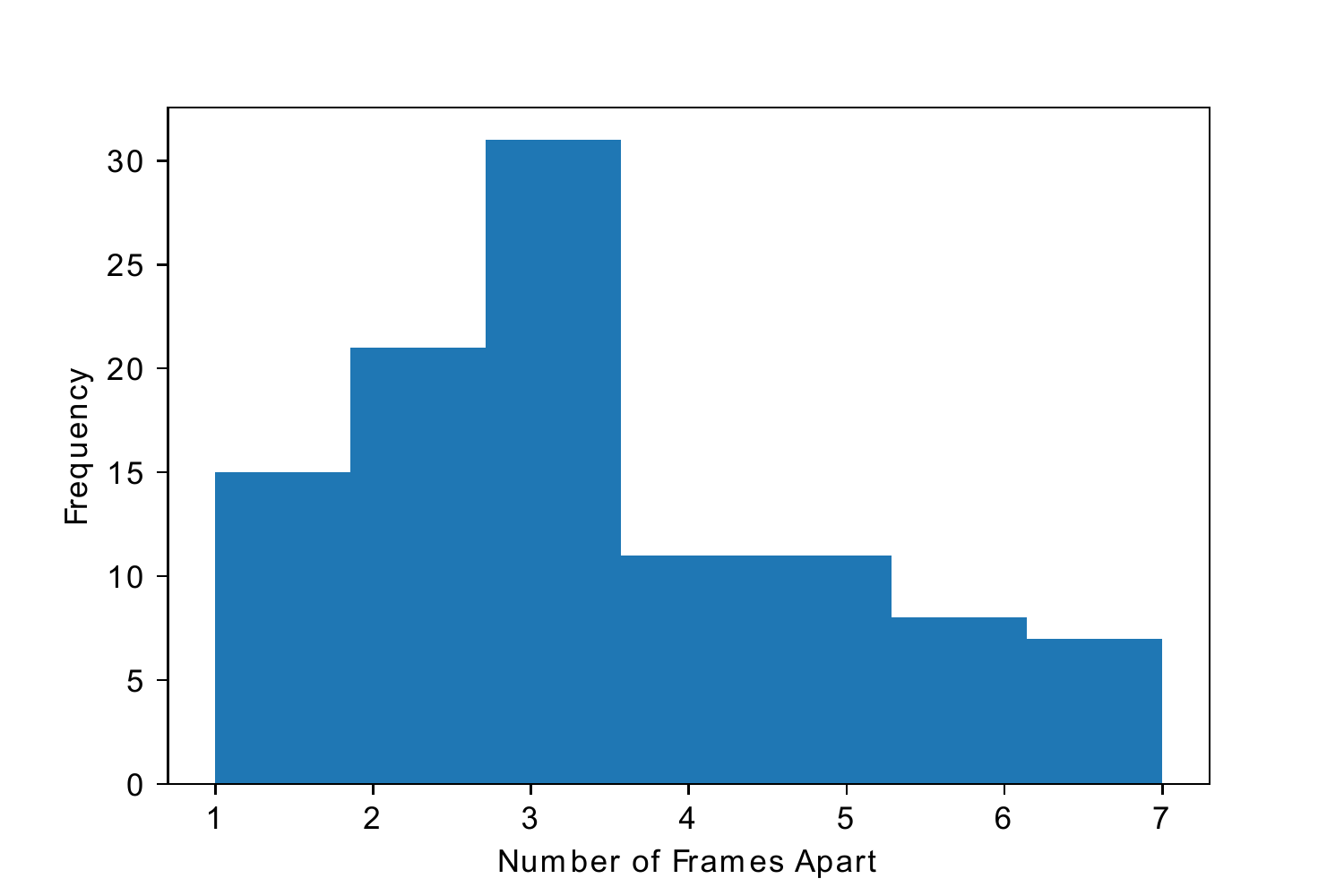}
    \caption{Distribution of the number of nodes apart for frame shifts.}
    \label{fig:num_nodes_apart_dist}
\end{figure}

\subsection{Quantitative Analysis of Frame Shifts}

\begin{figure}[htp]
    \centering
    \includegraphics[width=7cm]{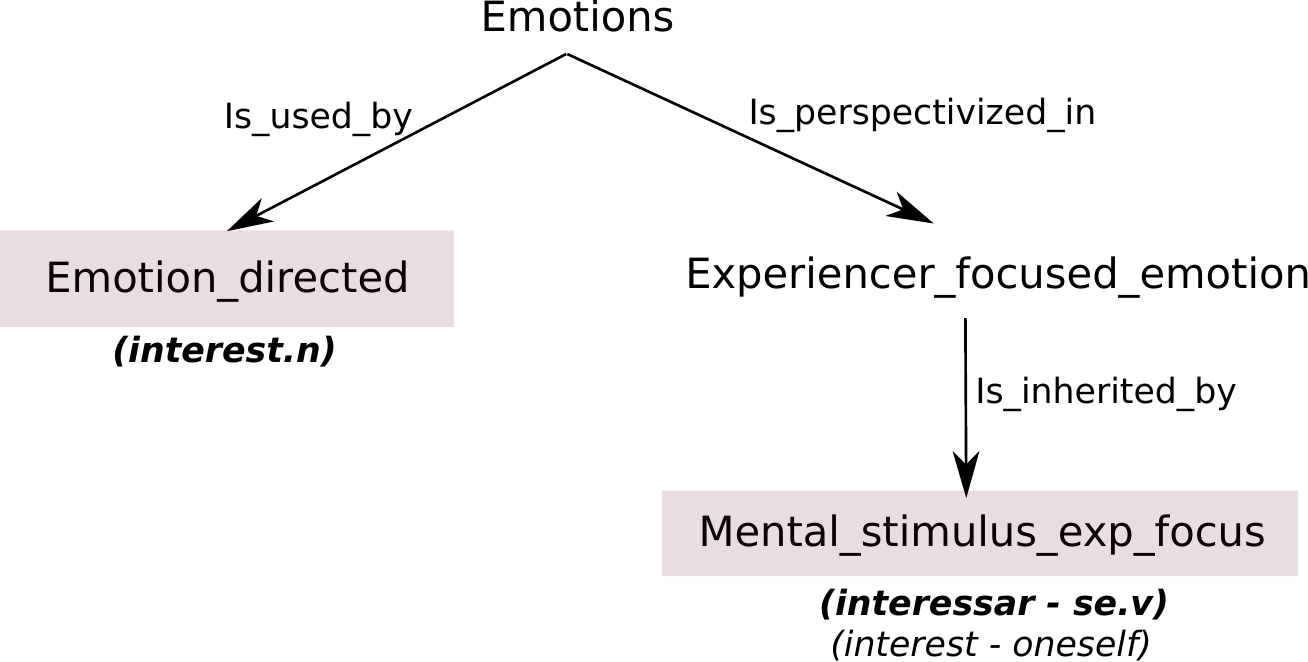}
    \caption{Visualization of the path connecting the pair of frames (\texttt{Emotion\_directed} and \texttt{Mental\_stimulus\_exp\_focus}) in frame shift in Figure~\ref{fig:frame_shift}.}
    \label{fig:frames_apart}
\end{figure}

Figure~\ref{fig:num_nodes_apart_dist} demonstrates a unimodal distribution of the distance in the FN network of diverging frames. Diverging frames do not necessarily exhibit \textit{first-order} frame-to-frame relations; many are more than one hop away from each other (see Figure~\ref{fig:frames_apart}). Most of the frame pairs are connected to each other. Even though not the full potential of connections FN is exploited to date, only two out of the 104 pairs of diverging frames do not have a path connecting them. In other words, frame shifts can be accounted for by the net-like configuration of FN, which is similar to the conclusion drawn by \citet{torrent2018multilingual}.

\section{Frame Shift Prediction (FSP)}
\subsection{Task Description}
FSP is a multi-class classification task. Given a labeled frame $f_{src} \in \mathcal{F}$, where $\mathcal{F}$ denotes the set of all 1224 frames in BFN 1.7 \cite{ruppenhofer2016framenet}, for a lexical unit ${LU}_{src}$ in the source English sentence, our goal is to predict the frame $f_{tgt} \in \mathcal{F}$ for the corresponding lexical unit ${LU}_{tgt}$ in the target German and Brazilian Portuguese sentences. %Following the Global FrameNet Shared Annotation Task \cite{torrent2018multilingual}, we use the same $\mathcal{F}$ for all three languages during training and inference.

\begin{figure}[htp]
    \centering
    \includegraphics[width=6cm , height=10cm]{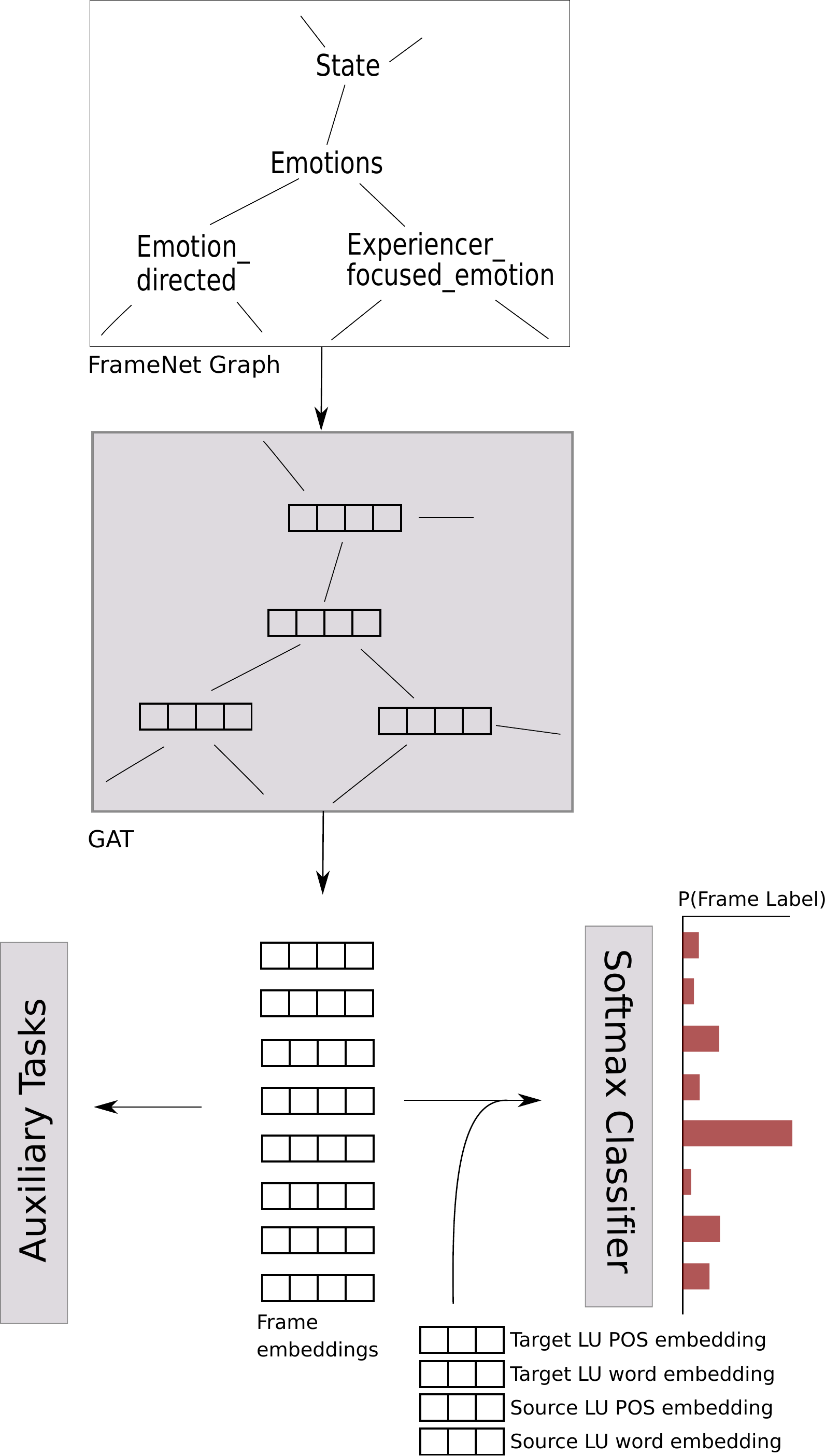}
    \caption{Proposed approach for FSP using graph attention networks (GAT) and auxiliary training to represent frames.}
    \label{fig:proposed_approach}
\end{figure}

\subsection{Proposed Model}
Figure~\ref{fig:proposed_approach} shows our proposed approach to FSP. We propose using graph attention networks (GATs) \cite{velivckovic2018graph} to represent frames to capture the relational structure of FN. The attention mechanism in GATs learns the weights of neighboring nodes according to node similarity and improves semantic clustering of frames \cite{ijcai2019graphclustering}.

% Table for Datasets
\begin{table*}[htp]
\centering
\begin{tabular}{lccccc}
\Xhline{2\arrayrulewidth}
\hline \textbf{Data} & \textbf{\# lus} & \textbf{\# frames} & \textbf{\# sents} & \textbf{langs} & \textbf{tasks}\\ \hline

\makecell[l]{Semantic Frame Shift\\ Prediction Dataset\\ (Section~\ref{sec:frame-shifts-dataset})} & 952 & 179 & 788 & \makecell{de, en, pt} & \makecell{Frame Shift Prediction} \\
\hline
\makecell[l]{Berkeley FrameNet 1.7 \\\cite{ruppenhofer2016framenet}}& 8404 & 1224 & 174527 & en & \makecell{Link Prediction \\ Path Length Prediction \\ Binary Frame Prediction \\ Frame Label Reconstruction} \\
\hline
\makecell[l]{Multilingual frame-\\annotated corpus\\ \cite{johannsen-martinezalonso-sogaard:2015:EMNLP}} & 7558 & 729 & 18442 & \makecell{bg, da, de, \\el, en, es, \\ fr, it, sv} & \makecell{Binary Frame Prediction \\ Frame Label Reconstruction} \\
\Xhline{2\arrayrulewidth}
\end{tabular}
\caption{\label{tab:dataset} Statistics of the all datasets in training GAT for FSP.}
\end{table*}

\subsubsection{Graph Initialization}
We define a graph $\mathcal{G} = (V, E, H)$ composed of a set of graph nodes $V$, node representations $H = (h_1, ..., h_{|V|})$ and a set of directed edges $E = (E_1, ..., E_K)$ where $K$ is the number of edges. Each of the 1224 nodes corresponds to a frame in $\mathcal{F}$. Each directed edge represents a frame-to-frame relation, and we do not distinguish between their types since, for the purposes of this study, it does not matter whether an edge captures a generic-specific relation (such as $Inheritance$) or causative-stative relation (such as $Causative\_of$), as both can be due to differences on how source and target language encode meaning. We initialize the nodes with multilingual LASER sentence representations of the frame definitions, where sentences from different languages are mapped into the same embedding space through a BiLSTM encoder \cite{artetxe2019massively}.

\subsubsection{Graph Attention Network (GAT)}
GAT consists of a stack of graph attentional layers \cite{velivckovic2018graph}. Each layer applies multi-headed self-attention mechanism to transform its inputs, which is a set of node representations $H = (h_1, ..., h_{|V|}), h_i \in \mathbb{R}^D$ (where $|V|$ is the number of nodes, and $D$ is the dimension of the node representation), to a new set of node representations, $H' = (h'_1, ..., h'_{|V|}), h'_i \in \mathbb{R}^{D'}$, of potentially different dimension $D'$. For $m$-th attention head, the output feature for a node $i$ is the linear combination of the input features of the node's first-order neighbors $j$ (including itself $i$), weighted by the normalized attention coefficients $\alpha^{(m)}_{ij}$. Then, the output features (which may undergo non-linear transformation $\sigma$) from all attention heads are concatenated to produce $h'_i$. The transformation from $h_i$ to $h'_i$ is as follows.
$$e^{(m)}_{ij} = \textrm{LeakyRELU}(\textbf{a}^{(m)T}[\textbf{W}^{(m)}h_i\|\textbf{W}^{(m)}h_j])$$
$$\alpha^{(m)}_{ij} = \frac{\exp(e^{(m)}_{ij})}{\sum_{k \in \mathcal{N}_i} \exp(e^{(m)}_{ik})}$$
$$h'_i = \overunderset{M}{m=1}\parallel \sigma \left( \sum_{j \in \mathcal{N}_i} \alpha^{(m)}_{ij} \textbf{W}^{(m)} h_j \right)$$

We follow the basic architecture of \citet{velivckovic2018graph} and implement a two-layer GAT model. Differing from \citet{velivckovic2018graph}, the output layer is a linear transformation layer followed by the softmax layer to include the LUs when the model calculates the probability for each frame $f_{tgt} \in \mathcal{F}$. The probability distribution $P(f_{tgt}|f_{src}, LU_{src}, LU_{tgt})$ is computed as follows: $${\rm softmax}(W([h_{f_{src}}; w_{src}; w_{tgt}; p_{src}; p_{tgt}]) + b)$$ where $h_{f_{src}}$ is the node representation of $f_{src}$; $w_{src}$ and $w_{tgt}$ are the pre-trained mBERT embeddings of $LU_{src}$ and $LU_{tgt}$ respectively; $p_{src}$ and $p_{tgt}$ are randomly initialized embeddings for parts-of-speech tags; $W$ and $b$ are the trainable parameters of the output layer. The whole system is trained with the cross-entropy loss function.

\subsubsection{Graph Regularization}
To improve the generalization ability of GAT and prevent it from overfitting, we further introduce two graph regularization techniques: NodeNorm \cite{zhou2020nodenorm} and DropEdge \cite{rong2019dropedge}. NodeNorm normalizes the embedding of every node in each layer by its own mean and standard deviation. It increases the smoothness of the model w.r.t. node features, so similar frames share similar representations. On the other hand, DropEdge randomly drops out a certain rate of edges of the input graph for each training time. As an unbiased data augmentation technique \cite{rong2019dropedge}, it enables a random subset aggregation instead of the full aggregation during GAT training, thus better capable of preventing overfitting. 

\begin{table*}[htp]
\centering
\begin{tabular}{lccc}
\Xhline{2\arrayrulewidth}
\textbf{Tasks} & \textbf{\# Layers} & \textbf{Layer Parameters} & \textbf{Objective Functions}\\ \hline
Frame Shift Prediction & 1 & $\mathbb{R}^{D_f + 2\times (D_w + D_{pos})}$ & Cross-Entropy Loss \\
Link Prediction & 1 & $\mathbb{R}^{2 \times D_f} \rightarrow \mathbb{R}^{2}$ & Cross-Entropy Loss \\
Path Length Prediction & 2 & $\mathbb{R}^{2 \times D_f} \rightarrow \mathbb{R}^{1024} \rightarrow \mathbb{R}$ & Mean-Squared Error \\
Binary Frame Prediction & 1 & $\mathbb{R}^{D_f + D_w + D_{pos}} \rightarrow \mathbb{R}^{2}$ & Cross-Entropy Loss \\
Frame Label Reconstruction & 1 & $\mathbb{R}^{D_f + 2\times (D_w + D_{pos})} \rightarrow \mathbb{R}^{1224}$ & Cross-Entropy Loss \\
\Xhline{2\arrayrulewidth}
\end{tabular}
\caption{\label{tab:model-parameters-aux-tasks} Parameters of output layers for frame shift prediction and auxiliary tasks.} %We use ReLU non-linearity in between the two output layers for the Path Length Prediction task (*).}
\end{table*}

\subsection{Auxiliary Training}
\label{sec:auxiliary-tasks}
We propose several auxiliary training tasks for GAT for two reasons. First, tasks 1 and 2 train our model to explicitly learn the connections among frames since we observe that 102 out of 104 pairs of diverging frames in our dataset are connected in FN. On the other hand, tasks 3 and 4 help our GAT associate frames with LUs since the relational structure of FN does not inform GAT how LUs evoke frames.

\begin{enumerate}
  \item \label{itm:aux-task-1} \textbf{Link Prediction.} A binary classification problem where the model predicts if there is a frame-to-frame relation between two semantic frames, $(f_1, f_2) \in E$.
  \item \label{itm:aux-task-2} \textbf{Path Length Prediction.} A regression task where the model predicts the number of edges between two frames, $(f_1, f_2) \in E$. 
  \item \label{itm:aux-task-3} \textbf{Binary Frame Prediction.} A binary classification task where, given a pair of randomly chosen frame $f$ and an LU $LU_x$, the model predicts if $LU_x$ evokes $f$.

  \item \label{itm:aux-task-4} \textbf{Frame Label Reconstruction.} A multi-class classification task where some of the frame labels $f$ for annotated sentences are randomly "perturbed" into incorrect frame labels $f_x$ with a probability $p$, and the model is trained to recover the correct frame $f$.
\end{enumerate}

Task 2 uses the mean squared error as the objective function whereas the rest uses cross entropy loss. The combined loss for training GAT is the sum of losses from the auxiliary tasks and the primary FSP task, weighted by the homoscedastic uncertainty of each task \cite{kendall2018multi}. 

\subsection{Datasets}
Table~\ref{tab:dataset} shows the statistics of the datasets used for FSP (primary task) and the auxiliary tasks. FSP experiments use the frame shifts dataset described in Section~\ref{sec:frame-shifts-dataset}. On the other hand, auxiliary tasks 1 and 2 use the frame-to-frame relationships information in BFN 1.7 \cite{ruppenhofer2016framenet} for training, whereas tasks 3 and 4 use the lexicographic annotations for the LUs in BFN 1.7 and the multilingual frame-annotated corpus \cite{johannsen-martinezalonso-sogaard:2015:EMNLP}.

\subsection{Experimental Setup}
We selected hyperparameters for GAT via Bayesian optimization on the Frame Label Reconstruction task and used the same hyperparameters for FSP. The resulting first layer of GAT consists of 9 attention heads computing 109 features each, and the second layer 10 attention heads 256 features each. The final softmax classifier only has a single linear transformation layer that receives 1824 input features from GAT and lexical units and outputs 1224 features (as frame classes). 

Table~\ref{tab:model-parameters-aux-tasks} shows the details of the final output layers for FSP and auxiliary tasks. We optimize their hyperparameters, namely the number of layers and the hidden features' dimension, on the respective auxiliary tasks before reusing hyperparameters for FSP. Here, we use $D_f = 256$ to denote the dimension of frame representations, $D_{pos} = 16$ the dimension of POS tag embeddings, and $D_{w} = 768$ the dimension of mBERT embeddings of LUs. 

We represent the LUs with mBERT embeddings \cite{devlin-etal-2019-bert}. Parts-of-speech tags are represented with randomly initialized embeddings of dimension 16. We train the model using the Adam optimizer with a batch size of 512, learning rate of 0.005, and weight decay of $\lambda=0.0005$. For each setting, we perform five runs of nested five-fold cross-validation on a Nvidia Tesla P100 GPU and report their average F1 scores as well as their standard deviations. The inner cross-validation is used to find the suitable number of training epochs. Training and evaluation model take approximately three hours.

%%%%% Table : Frame Shift Prediction Dataset
\begin{table*}[htp]
\centering
\begin{tabular}{lccc}
\Xhline{2\arrayrulewidth}
\textbf{Models} & \textbf{EN $\rightarrow$ PT} & \textbf{EN $\rightarrow$ DE} & \textbf{EN $\rightarrow$ (PT + DE)}\\ \hline
Direct Transfer & 77.5 & 63.9 & 74.3 \\
Randomized Embeddings & 44.2 ($\pm$ 3.1) & 37.7 ($\pm$ 2.4) & 39.8 ($\pm$ 2.9) \\
\citet{sikos-pado-2018-framenet-comparison} & - & 55.2 ($\pm$ 3.5) & - \\
mBERT (w/o fine-tuning) \cite{sikos-pado-2019-frame-exemplars-prototypes} & 53.4 ($\pm$ 1.7) & 47.8 ($\pm$ 4.3) & 49.3 ($\pm$ 2.4) \\
mBERT (with fine-tuning) \cite{sikos-pado-2019-frame-exemplars-prototypes} & 71.5 ($\pm$ 2.3) & 65.6 ($\pm$ 0.9) & 68.7 ($\pm$ 2.2) \\
FastText \cite{baker-lorenzi-2020-exploring} & 54.8 ($\pm$ 1.1) & 43.7 ($\pm$ 2.6) & 50.1 ($\pm$ 1.8) \\ 
\hline
GAT (w/o auxiliary training) & 57.1 ($\pm$ 1.3) & 40.2 ($\pm$ 1.8) & 55.9  ($\pm$ 4.1) \\
\textbf{GAT (with auxiliary training)} & \textbf{83.1} ($\pm$ 1.5) & \textbf{68.0} ($\pm$ 1.9) & \textbf{79.7} ($\pm$ 2.0) \\
\Xhline{2\arrayrulewidth}
\end{tabular}
\caption{\label{tab:frame-shift-pred-result} 5-Fold nested cross-validation with top-5 F1 scores ($\pm$ standard deviation) for each model in predicting frame shifts. $X \rightarrow Y$ denotes that projecting frames from language $X$ to language $Y$ (EN: English, PT: Portuguese, DE: German).}
\end{table*}

\subsection{Baselines}
Since our paper is the first attempt to predict frame shifts, we do not have other classifiers to directly compare with. As we are proposing a novel frame representation method for the multilingual task, we use other recent multilingual frame representation methods as baselines. The frame representations obtained from the baselines are concatenated with the word embeddings and part-of-speech tag embeddings of the LUs. Subsequently, the tensors are passed through a single linear transformation layer and a softmax final layer for classification.

\paragraph{Direct Transfer.} This method assumes that frame shifts are absent and projects the frame labels without changes. In other words, $f_{tgt} = f_{src}$.

\paragraph{Randomized Frame Embeddings.} This method represents each frame with a trainable, randomized embedding of dimension 256. The embeddings are cross-lingual because they are trained with the FSP dataset.

\paragraph{\citet{sikos-pado-2018-framenet-comparison}.} The authors embedded English and German frames from BFN 1.5 and SALSA corpus in the same vector space. In our setup, the frame embeddings are only used for FSP between English and German. We directly transfer the frames that are not embedded by the authors.

\paragraph{\citet{sikos-pado-2019-frame-exemplars-prototypes}.} The authors embedded frames with the pre-trained BERT model with and without fine-tuning. Without fine-tuning, frame embeddings are the unweighted centroid of the contextualized embeddings of the corresponding LUs. Otherwise, the frame embeddings are fine-tuned to predict frame labels for each word token in the full-text annotations. In our setup, we use the multilingual BERT (mBERT) model to represent frames and fine-tune them on all the datasets in Table~\ref{tab:dataset} following the authors' instructions.

\paragraph{\citet{baker-lorenzi-2020-exploring}.} The authors created frame embeddings using the unweighted centroid of the FastText embeddings of the LUs. In our experiments, we embed the LUs in the source and target sentences with FastText representations.

% On top of those three models, we also compare our model to: (a) one using direct transfer, i.e., assuming that frame shifts are absent and projecting the frame labels without; (b) another one using randomized frame embeddings, i.e., representing each frame with a trainable, randomized embedding of dimension 256. % The embeddings are cross-lingual because they are trained with the FSP dataset.

\subsection{Evaluation}
To ensure a robust evaluation of models on a small FSP dataset, we evaluate each model with the five-fold nested cross-validation (CV) method. It separates the CV fold used for model development (including feature selection and parameter tuning) from the one used for model evaluation; therefore, the performance estimates are unaffected by and unbiased to the sample sizes \cite{vabalas2019machine}. 

We evaluate FSP with the top-5 F1 score. As long as the correct frame label is among the top-5 most probable predicted frame shifts–––hence the term "top-5"–––we consider the model to have successfully predicted
the frame shift. The reason for this metric choice is that the size of our FSP dataset is much smaller than the number of classes (1224 frames with varying granularity) in this experiment. As a result, the models have to perform FSP on frames they have not seen before in FSP training. Furthermore, the frame labels vary with respect to granularity, which can cause the model to suffer from class ambiguity. Therefore, the top-5 F1 score gives a more realistic performance evaluation.

\section{Discussion}

%%%%% Figure : UMAP Representations
\begin{figure*}[htp]
    \centering
    \includegraphics[width=10.5cm , height=4.5cm]{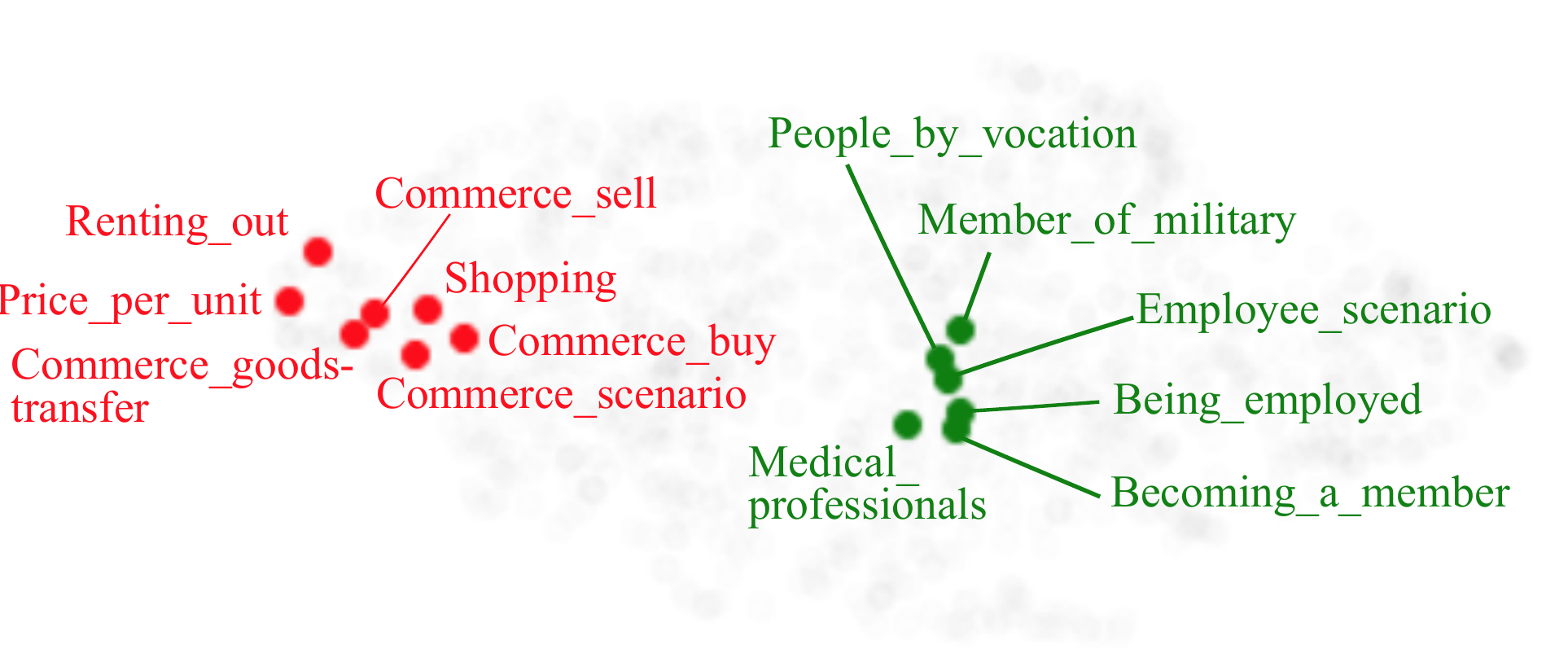}
    \caption{UMAP visualization of semantic frame vectors learned by our proposed graph attention networks model. We color-code the frames related to commerce (in red) and occupation (in green) to show the clustering of frames.}
    \label{fig:umap_semantic_frame}
\end{figure*}

\subsection{Frame Shift Prediction}
Table~\ref{tab:frame-shift-pred-result} illustrates the performance of different models in FSP. Embedding-based baseline models generally perform worse than the Direct Tansfer approach. Hence, we argue that simply aggregating pre-trained contextualized representations of LUs to represent frames cannot capture the fine-grained semantic distinctions between frames. One solution is to fine-tune the frame representations on the frame induction task. It encodes the similarities and variations between frames and better separates the frames that involve polysemous lemmas evoking multiple frames \cite{sikos-pado-2019-frame-exemplars-prototypes}. As seen in Table~\ref{tab:frame-shift-pred-result}, the fine-tuning approach reports the best performance among the embedding-based baselines.

Our use of GAT and auxiliary training to represent frames achieves the best performance. We want to highlight the differences in our approach: our model treats frames and LUs as independent units and learns their relations through the auxiliary tasks~\ref{itm:aux-task-3} and~\ref{itm:aux-task-4}, as opposed to representing frames as a combination of LU representations \cite{peng-etal-2018-learning,sikos-pado-2018-framenet-comparison,sikos-pado-2019-frame-exemplars-prototypes,popov-sikos-2019-graph,alhoshan-etal-2019-semantic,baker-lorenzi-2020-exploring}. Furthermore, our approach can represent non-lexical frames because of the message propagation from their surrounding nodes. The main takeaway here is that learning the relational structure of FN enables FSP. %Furthermore, our approach can represent non-lexical frames because of the message propagation from their surrounding nodes. 

We see a significant decrease in performance when auxiliary training is absent. Auxiliary training boosts performance for the FSP in Brazilian Portuguese (EN $\rightarrow$ PT) even though the auxiliary datasets (see Table~\ref{tab:dataset}) do not contain Brazilian Portuguese sentences. This shows that the auxiliary learning successfully encodes cross-linguistic information about frames --- it helps the GAT model to learn the frame-to-frame and LU-to-frame relationships, thus creating more generalized and meaningful frame representations.

The study limitation is the small sample size. Global FrameNet \cite{torrent2018multilingual} is an initiative involving several languages, and the shared annotation task requires fine grained annotation of the parallel corpora. Nonetheless, it is the dataset the FN community currently uses for studying frame adequacy across languages.

\subsection{Visualization of Frame Representations}

Figure~\ref{fig:umap_semantic_frame} illustrates the semantic frame representations with UMAP dimensionality reduction \cite{mcinnes2018umap}. To obtain the frame representations, we average the node representations from five different GAT models trained in the five-fold nested cross-validation. We foreground two clear clusters of frames, where one is related to commerce and the other related to occupation, to show that GAT learns the underlying relationships between frames. 

There are two main findings. First, the frames in the clusters are connected to one another in FN; for instance, \texttt{Member\_of\_military} inherits from \texttt{People\_by\_vocation}, and \texttt{Being\_employed} is a subframe of \texttt{Employee\_scenario}. Second, we obtain similar representations for frames that share a domain but are not connected. \texttt{Price\_per\_unit} and \texttt{Commercial\_goods-transfer} are both conceptually related to commerce, and there is no path connecting them in FN, but they are still clustered together because of their semantic associations. The clustering is a clear sign that the GAT model has successfully learned the relationships between frames. 

\subsection{Ablation Study}
%%%% Table - Ablation Study
\begin{table}[htp]
\centering
\begin{tabular}{lcc}
\Xhline{2\arrayrulewidth}
\textbf{Models} & \textbf{F1} & \textbf{$\Delta$}\\ \hline
GAT (all Auxiliary Tasks) & 79.7 & -\\
-- Link Prediction & 76.1 & -3.6\\
-- Path Length Prediction & 75.5 & -4.2\\
-- Binary Frame Prediction & 69.4 & -10.3\\
-- Frame Label Reconstruction  & 63.7 & -16.0\\
-- All Auxiliary Tasks & 55.9 & -23.8\\
\Xhline{2\arrayrulewidth}
\end{tabular}
\caption{\label{tab:ablation} Ablation study on auxiliary tasks.}
\end{table}

Table~\ref{tab:ablation} shows the result of an ablation study of auxiliary tasks. We conclude that the auxiliary tasks are suitable for learning frame shifts as ablation of any task hurts FSP. Out of the four tasks, Frame Label Reconstruction is the most helpful for learning FSP. This could be due to shared task structure and classifier parameters between the auxiliary task and the FSP task, as both tasks compute the posterior probability of frame labels for a LU given a prior (source) frame. In contrast, Link Prediction contributes the least to FSP. This is possibly due the presence of frames that are not immediate neighbors in FrameNet, making supervised learning of frame-to-frame relations less informative. 

\section{Conclusion and Future Work}
Our research details the causes of frame shifts on the morpho-syntactic, structural, and construal levels. We also pioneer a new task, Frame Shift Prediction (FSP), and show that Graph Attention Networks (GATs) can predict frame shifts by learning FrameNet's relational structure and the lexical units. In the future, we plan to explore automatically categorizing frame shifts according to potential factors to improve FSP performance.

\section*{Ethical Considerations}
In this paper, we present the Frame Shift Prediction dataset that is created automatically from the Global FrameNet Shared Annotation Task \cite{torrent2018multilingual}, as detailed in Section~\ref{sec:frame-shifts-dataset}. The curation is consistent with the terms, the intellectual property and privacy rights of the original dataset. FrameNet-like annotation, as the one performed in the Global FrameNet Shared Annotation Task, is carried-out by highly trained linguists and uses fine-grained culturally grounded labels that are proposed based on the study of balanced corpora, such as the British National Corpus and the American National Corpus, for English. 

Our research enables the automatic bootstrapping of multilingual FrameNets, which could improve semantics-based NLP applications including machine translation, question-answering, and information retrieval systems. These applications may have greater reach due to their multilinguality and contribute to a large range of services, such as helpdesks, personal assistants, retail and sales. On the other hand, newly created FrameNets that rely exclusively in the transposition of structure from another language may present source-language induced biases, which can be corrected through the validation of the bootstrapped structure against corpus evidence in the target language.

\section*{Acknowledgements}
This material is based in part upon work supported by the U.S. National Science Foundation under Grant No. (1629989). Any opinions, findings, and conclusions or recommendations expressed in this material are those of the author(s) and do not necessarily reflect the views of the National Science Foundation. The annotation of the TED Talk for Brazilian Portuguese and German were made possible by the CAPES/PROBRAL Grant No. 88881.144044/2017-00.

\bibliography{ori_anthology,custom}

\begin{thebibliography}{46}
\expandafter\ifx\csname natexlab\endcsname\relax\def\natexlab#1{#1}\fi

\bibitem[{Alhoshan et~al.(2019)Alhoshan, Batista-Navarro, and
  Zhao}]{alhoshan-etal-2019-semantic}
Waad Alhoshan, Riza Batista-Navarro, and Liping Zhao. 2019.
\newblock \href {https://doi.org/10.18653/v1/W19-0606} {Semantic frame
  embeddings for detecting relations between software requirements}.
\newblock In \emph{Proceedings of the 13th International Conference on
  Computational Semantics - Student Papers}, pages 44--51, Gothenburg, Sweden.
  Association for Computational Linguistics.

\bibitem[{Artetxe and Schwenk(2019)}]{artetxe2019massively}
Mikel Artetxe and Holger Schwenk. 2019.
\newblock \href {https://transacl.org/ojs/index.php/tacl/article/view/1742}
  {Massively multilingual sentence embeddings for zero-shot cross-lingual
  transfer and beyond}.
\newblock \emph{Transactions of the Association for Computational Linguistics},
  7:597--610.

\bibitem[{Baker and Lorenzi(2020)}]{baker-lorenzi-2020-exploring}
Collin~F. Baker and Arthur Lorenzi. 2020.
\newblock \href {https://www.aclweb.org/anthology/2020.framenet-1.11}
  {Exploring crosslinguistic frame alignment}.
\newblock In \emph{Proceedings of the International FrameNet Workshop 2020:
  Towards a Global, Multilingual FrameNet}, pages 77--84, Marseille, France.
  European Language Resources Association.

\bibitem[{Boas and Ziem(2018)}]{boas2018constructing}
Hans~C Boas and Alexander Ziem. 2018.
\newblock \href {https://benjamins.com/catalog/cal.22.07boa} {Constructing a
  constructicon for german}.
\newblock \emph{Constructicography: Constructicon development across
  languages}, 22:183.

\bibitem[{Bogin et~al.(2019)Bogin, Berant, and
  Gardner}]{bogin-etal-2019-representing}
Ben Bogin, Jonathan Berant, and Matt Gardner. 2019.
\newblock \href {https://doi.org/10.18653/v1/P19-1448} {Representing schema
  structure with graph neural networks for text-to-{SQL} parsing}.
\newblock In \emph{Proceedings of the 57th Annual Meeting of the Association
  for Computational Linguistics}, pages 4560--4565, Florence, Italy.
  Association for Computational Linguistics.

\bibitem[{Bond and Paik(2012)}]{Bond:Paik:2012}
Francis Bond and Kyonghee Paik. 2012.
\newblock \href
  {https://pdfs.semanticscholar.org/1169/e0d43ac805271600c27f8bc59a2137b1ff3f.pdf}
  {{A} {S}urvey of {W}ord{N}ets and their {L}icenses}.
\newblock In \emph{Proceedings of the 6th Global WordNet Conference (GWC
  2012)}, Matsue.
\newblock 64--71.

\bibitem[{Burchardt et~al.(2006)Burchardt, Erk, Frank, Kowalski, Pad{\'o}, and
  Pinkal}]{burchardt-etal-2006-salsa}
Aljoscha Burchardt, Katrin Erk, Anette Frank, Andrea Kowalski, Sebastian
  Pad{\'o}, and Manfred Pinkal. 2006.
\newblock \href {http://www.lrec-conf.org/proceedings/lrec2006/pdf/339_pdf.pdf}
  {The {SALSA} corpus: a {G}erman corpus resource for lexical semantics}.
\newblock In \emph{Proceedings of the Fifth International Conference on
  Language Resources and Evaluation ({LREC}{'}06)}, Genoa, Italy. European
  Language Resources Association (ELRA).

\bibitem[{{\v{C}}ulo(2013)}]{vculo2013constructions}
Oliver {\v{C}}ulo. 2013.
\newblock \href {https://doi.org/10.1075/cf.5.2.02cul}
  {Constructions-and-frames analysis of translations: {T}he interplay of syntax
  and semantics in translations between {E}nglish and {G}erman}.
\newblock \emph{Constructions and Frames}, 5(2):143--167.

\bibitem[{Czulo(2017)}]{czulo2017aspects}
Oliver Czulo. 2017.
\newblock Aspects of a primacy of frame model of translation.
\newblock In S.~Hansen-Schirra, Oliver Czulo, and Sascha Hofmann, editors,
  \emph{Empirical modelling of translation and interpreting}, number~6 in
  Translation and {Multilingual} {Natural} {Language} {Processing}, pages
  465--490. Language Science Press, Berlin.

\bibitem[{Devlin et~al.(2019)Devlin, Chang, Lee, and
  Toutanova}]{devlin-etal-2019-bert}
Jacob Devlin, Ming-Wei Chang, Kenton Lee, and Kristina Toutanova. 2019.
\newblock \href {https://doi.org/10.18653/v1/N19-1423} {{BERT}: Pre-training of
  deep bidirectional transformers for language understanding}.
\newblock In \emph{Proceedings of the 2019 Conference of the North {A}merican
  Chapter of the Association for Computational Linguistics: Human Language
  Technologies, Volume 1 (Long and Short Papers)}, pages 4171--4186,
  Minneapolis, Minnesota. Association for Computational Linguistics.

\bibitem[{Dorr(1994)}]{dorr-1994-machine}
Bonnie~J. Dorr. 1994.
\newblock \href {https://www.aclweb.org/anthology/J94-4004} {Machine
  translation divergences: A formal description and proposed solution}.
\newblock \emph{Computational Linguistics}, 20(4):597--633.

\bibitem[{Dyer et~al.(2013)Dyer, Chahuneau, and Smith}]{dyer-etal-2013-simple}
Chris Dyer, Victor Chahuneau, and Noah~A. Smith. 2013.
\newblock \href {https://www.aclweb.org/anthology/N13-1073} {A simple, fast,
  and effective reparameterization of {IBM} model 2}.
\newblock In \emph{Proceedings of the 2013 Conference of the North {A}merican
  Chapter of the Association for Computational Linguistics: Human Language
  Technologies}, pages 644--648, Atlanta, Georgia. Association for
  Computational Linguistics.

\bibitem[{Fillmore(1982)}]{fillmore82frame}
Charles~J. Fillmore. 1982.
\newblock \emph{Frame semantics}, pages 111--137. Hanshin Publishing Co.,
  Seoul, South Korea.

\bibitem[{Gargett and Leung(2020)}]{gargett-leung-2020-building}
Andrew Gargett and Tommi Leung. 2020.
\newblock \href {https://www.aclweb.org/anthology/2020.framenet-1.10} {Building
  the emirati {A}rabic {F}rame{N}et}.
\newblock In \emph{Proceedings of the International FrameNet Workshop 2020:
  Towards a Global, Multilingual FrameNet}, pages 70--76, Marseille, France.
  European Language Resources Association.

\bibitem[{Gilardi and Baker(2018)}]{gilardi2018multilingual}
Luca Gilardi and Collin Baker. 2018.
\newblock \href {http://lrec-conf.org/workshops/lrec2018/W5/pdf/11_W5.pdf}
  {Learning to align across languages: Toward multilingual framenet}.
\newblock In \emph{Proceedings of the Eleventh International Conference on
  Language Resources and Evaluation (LREC 2018)}, Paris, France. European
  Language Resources Association (ELRA).

\bibitem[{Giouli et~al.(2020)Giouli, Pilitsidou, and
  Christopoulos}]{giouli-etal-2020-greek}
Voula Giouli, Vera Pilitsidou, and Hephaestion Christopoulos. 2020.
\newblock \href {https://www.aclweb.org/anthology/2020.framenet-1.7} {{G}reek
  within the global {F}rame{N}et initiative: Challenges and conclusions so
  far}.
\newblock In \emph{Proceedings of the International FrameNet Workshop 2020:
  Towards a Global, Multilingual FrameNet}, pages 48--55, Marseille, France.
  European Language Resources Association.

\bibitem[{Ji et~al.(2019)Ji, Wu, and Lan}]{ji-etal-2019-graph}
Tao Ji, Yuanbin Wu, and Man Lan. 2019.
\newblock \href {https://doi.org/10.18653/v1/P19-1237} {Graph-based dependency
  parsing with graph neural networks}.
\newblock In \emph{Proceedings of the 57th Annual Meeting of the Association
  for Computational Linguistics}, pages 2475--2485, Florence, Italy.
  Association for Computational Linguistics.

\bibitem[{Johannsen et~al.(2015)Johannsen, Mart\'{i}nez~Alonso, and
  S{\o}gaard}]{johannsen-martinezalonso-sogaard:2015:EMNLP}
Anders Johannsen, H\'{e}ctor Mart\'{i}nez~Alonso, and Anders S{\o}gaard. 2015.
\newblock \href {http://aclweb.org/anthology/D15-1245} {Any-language
  frame-semantic parsing}.
\newblock In \emph{Proceedings of the 2015 Conference on Empirical Methods in
  Natural Language Processing}, pages 2062--2066, Lisbon, Portugal. Association
  for Computational Linguistics.

\bibitem[{Johansson and Nugues(2006)}]{johansson-nugues-2006-framenet}
Richard Johansson and Pierre Nugues. 2006.
\newblock \href {https://www.aclweb.org/anthology/P06-2057} {A
  {F}rame{N}et-based semantic role labeler for {S}wedish}.
\newblock In \emph{Proceedings of the {COLING}/{ACL} 2006 Main Conference
  Poster Sessions}, pages 436--443, Sydney, Australia. Association for
  Computational Linguistics.

\bibitem[{Kendall et~al.(2018)Kendall, Gal, and Cipolla}]{kendall2018multi}
Alex Kendall, Yarin Gal, and Roberto Cipolla. 2018.
\newblock \href
  {https://openaccess.thecvf.com/content_cvpr_2018/papers/Kendall_Multi-Task_Learning_Using_CVPR_2018_paper.pdf}
  {Multi-task learning using uncertainty to weigh losses for scene geometry and
  semantics}.
\newblock In \emph{Proceedings of the IEEE conference on computer vision and
  pattern recognition}, pages 7482--7491.

\bibitem[{Li et~al.(2017)Li, Tapaswi, Liao, Jia, Urtasun, and
  Fidler}]{li2017situation}
Ruiyu Li, Makarand Tapaswi, Renjie Liao, Jiaya Jia, Raquel Urtasun, and Sanja
  Fidler. 2017.
\newblock Situation recognition with graph neural networks.
\newblock In \emph{Proceedings of the IEEE International Conference on Computer
  Vision}, pages 4173--4182.

\bibitem[{Lind{\'e}n et~al.(2019)Lind{\'e}n, Haltia, Laine, Luukkonen,
  Piitulainen, and V{\"a}is{\"a}nen}]{linden2019finntransframe}
Krister Lind{\'e}n, Heidi Haltia, Antti Laine, Juha Luukkonen, Jussi
  Piitulainen, and Niina V{\"a}is{\"a}nen. 2019.
\newblock \href {https://doi.org/10.1007/s10579-018-9434-y} {Finntransframe:
  translating frames in the finnframenet project}.
\newblock \emph{Language Resources and Evaluation}, 53(1):141--171.

\bibitem[{Litkowski(2009)}]{litkowski2009hans}
Kenneth~C Litkowski. 2009.
\newblock \href {https://doi.org/10.1515/9783110212976} {{M}ultilingual
  {F}ramenets in {C}omputational {L}exicography: {M}ethods and {A}pplications.}

\bibitem[{Marcheggiani and Titov(2017)}]{marcheggiani-titov-2017-encoding}
Diego Marcheggiani and Ivan Titov. 2017.
\newblock \href {https://doi.org/10.18653/v1/D17-1159} {Encoding sentences with
  graph convolutional networks for semantic role labeling}.
\newblock In \emph{Proceedings of the 2017 Conference on Empirical Methods in
  Natural Language Processing}, pages 1506--1515, Copenhagen, Denmark.
  Association for Computational Linguistics.

\bibitem[{McInnes et~al.(2018)McInnes, Healy, and Melville}]{mcinnes2018umap}
Leland McInnes, John Healy, and James Melville. 2018.
\newblock \href {https://doi.org/10.21105/joss.00861} {{UMAP}: {U}niform
  {M}anifold {A}pproximation and {P}rojection}.
\newblock \emph{Journal of Open Source Software}, 3(29):861.

\bibitem[{Nathani et~al.(2019)Nathani, Chauhan, Sharma, and
  Kaul}]{nathani-etal-2019-learning}
Deepak Nathani, Jatin Chauhan, Charu Sharma, and Manohar Kaul. 2019.
\newblock \href {https://doi.org/10.18653/v1/P19-1466} {Learning
  attention-based embeddings for relation prediction in knowledge graphs}.
\newblock In \emph{Proceedings of the 57th Annual Meeting of the Association
  for Computational Linguistics}, pages 4710--4723, Florence, Italy.
  Association for Computational Linguistics.

\bibitem[{Ohara(2020)}]{ohara-2020-finding}
Kyoko Ohara. 2020.
\newblock \href {https://www.aclweb.org/anthology/2020.framenet-1.2} {Finding
  corresponding constructions in {E}nglish and {J}apanese in a {TED} talk
  parallel corpus using frames-and-constructions analysis}.
\newblock In \emph{Proceedings of the International FrameNet Workshop 2020:
  Towards a Global, Multilingual FrameNet}, pages 8--12, Marseille, France.
  European Language Resources Association.

\bibitem[{Pad\'{o} and Lapata(2009)}]{pado2009cross}
Sebastian Pad\'{o} and Mirella Lapata. 2009.
\newblock \href {https://dl.acm.org/doi/10.5555/1734953.1734960}
  {{C}ross-{L}ingual {A}nnotation {P}rojection of {S}emantic roles}.
\newblock \emph{Journal of Artificial Intelligence Research}, 36(1).

\bibitem[{Peng et~al.(2018)Peng, Thomson, Swayamdipta, and
  Smith}]{peng-etal-2018-learning}
Hao Peng, Sam Thomson, Swabha Swayamdipta, and Noah~A. Smith. 2018.
\newblock \href {https://doi.org/10.18653/v1/N18-1135} {Learning joint semantic
  parsers from disjoint data}.
\newblock In \emph{Proceedings of the 2018 Conference of the North {A}merican
  Chapter of the Association for Computational Linguistics: Human Language
  Technologies, Volume 1 (Long Papers)}, pages 1492--1502, New Orleans,
  Louisiana. Association for Computational Linguistics.

\bibitem[{Popov and Sikos(2019)}]{popov-sikos-2019-graph}
Alexander Popov and Jennifer Sikos. 2019.
\newblock \href {https://doi.org/10.26615/978-954-452-056-4_109} {Graph
  embeddings for frame identification}.
\newblock In \emph{Proceedings of the International Conference on Recent
  Advances in Natural Language Processing (RANLP 2019)}, pages 939--948, Varna,
  Bulgaria. INCOMA Ltd.

\bibitem[{Rong et~al.(2020)Rong, Huang, Xu, and Huang}]{rong2019dropedge}
Yu~Rong, Wenbing Huang, Tingyang Xu, and Junzhou Huang. 2020.
\newblock \href {https://openreview.net/forum?id=Hkx1qkrKPr} {{D}rop{E}dge:
  {T}owards {D}eep {G}raph {C}onvolutional {N}etworks on {N}ode
  {C}lassification}.
\newblock In \emph{International Conference on Learning Representations}.

\bibitem[{Ruppenhofer et~al.(2016)Ruppenhofer, Ellsworth, Schwarzer-Petruck,
  Johnson, Baker, and Scheffczyk}]{ruppenhofer2016framenet}
Josef Ruppenhofer, Michael Ellsworth, Myriam Schwarzer-Petruck, Christopher~R
  Johnson, Collin~F. Baker, and Jan Scheffczyk. 2016.
\newblock \href {https://framenet2.icsi.berkeley.edu/docs/r1.7/book.pdf}
  {Frame{N}et {II}: {E}xtended {T}heory and {P}ractice}.
\newblock \emph{FrameNet Project}.

\bibitem[{Shang et~al.(2019)Shang, Tang, Huang, Bi, He, and
  Zhou}]{shang2019end}
Chao Shang, Yun Tang, Jing Huang, Jinbo Bi, Xiaodong He, and Bowen Zhou. 2019.
\newblock End-to-end structure-aware convolutional networks for knowledge base
  completion.
\newblock In \emph{Proceedings of the AAAI Conference on Artificial
  Intelligence}, volume~33, pages 3060--3067.

\bibitem[{Sikos and Pad{\'o}(2018)}]{sikos-pado-2018-framenet-comparison}
Jennifer Sikos and Sebastian Pad{\'o}. 2018.
\newblock \href {https://www.aclweb.org/anthology/W18-3813} {Using embeddings
  to compare {F}rame{N}et frames across languages}.
\newblock In \emph{Proceedings of the First Workshop on Linguistic Resources
  for Natural Language Processing}, pages 91--101, Santa Fe, New Mexico, USA.
  Association for Computational Linguistics.

\bibitem[{Sikos and
  Pad{\'o}(2019)}]{sikos-pado-2019-frame-exemplars-prototypes}
Jennifer Sikos and Sebastian Pad{\'o}. 2019.
\newblock \href {https://doi.org/10.18653/v1/W19-0425} {Frame identification as
  categorization: Exemplars vs prototypes in embeddingland}.
\newblock In \emph{Proceedings of the 13th International Conference on
  Computational Semantics - Long Papers}, pages 295--306, Gothenburg, Sweden.
  Association for Computational Linguistics.

\bibitem[{Subirats and Sato(2003)}]{subirats2003surprise}
Carlos Subirats and Hiroaki Sato. 2003.
\newblock \href
  {https://doi.org/http://citeseerx.ist.psu.edu/viewdoc/summary?doi=10.1.1.585.1620}
  {{S}urprise! {S}panish {F}rame{N}et}.
\newblock In \emph{Proceedings of the Workshop on Frame Semantics at the XVII.
  International Congress of Linguists}. Citeseer.

\bibitem[{Suhail and Sigal(2019)}]{suhail2019mixture}
Mohammed Suhail and Leonid Sigal. 2019.
\newblock Mixture-kernel graph attention network for situation recognition.
\newblock In \emph{Proceedings of the IEEE International Conference on Computer
  Vision}, pages 10363--10372.

\bibitem[{Torrent and Ellsworth(2013)}]{torrent2013behind}
Tiago~Timponi Torrent and Michael Ellsworth. 2013.
\newblock \href
  {https://periodicos.ufjf.br/index.php/veredas/article/view/25403} {Behind the
  labels: criteria for defining analytical categories in framenet brasil}.
\newblock \emph{Veredas-Revista de Estudos Linguisticos}, 17(1):44--66.

\bibitem[{Torrent et~al.(2018{\natexlab{a}})Torrent, Ellsworth, Baker, and
  Matos}]{torrent2018multilingual}
Tiago~Timponi Torrent, Michael Ellsworth, Collin Baker, and Ely Edison da~Silva
  Matos. 2018{\natexlab{a}}.
\newblock \href {http://lrec-conf.org/workshops/lrec2018/W5/pdf/12_W5.pdf}
  {{T}he {M}ultilingual {F}rame{N}et {S}hared {A}nnotation {T}ask: a
  {P}reliminary {R}eport}.
\newblock In \emph{Proceedings of the Eleventh International Conference on
  Language Resources and Evaluation (LREC 2018)}, Paris, France. European
  Language Resources Association (ELRA).

\bibitem[{Torrent et~al.(2018{\natexlab{b}})Torrent, Matos, Lage, Laviola,
  Tavares, Almeida, and Sigiliano}]{torrent2018fnbr}
Tiago~Timponi Torrent, Ely Edison da~Silva Matos, Ludmila Lage, Adrieli
  Laviola, Tatiane Tavares, Vânia Gomes~de Almeida, and Natália Sigiliano.
  2018{\natexlab{b}}.
\newblock \href {https://benjamins.com/catalog/cal.22.04tor} {Towards
  continuity between the lexicon and the constructicon in framenet brasil}.
\newblock In Benjamin Lyngfelt, Lars Borin, Kyoko Ohara, and Tiago~Timponi
  Torrent, editors, \emph{Constructicography: constructicon development across
  languages}, pages 107--140. John Benjamins, Amsterdam, The Netherlands.

\bibitem[{Trott et~al.(2020)Trott, Torrent, Chang, and
  Schneider}]{trott-etal-2020-construing}
Sean Trott, Tiago~Timponi Torrent, Nancy Chang, and Nathan Schneider. 2020.
\newblock \href {https://doi.org/10.18653/v1/2020.acl-main.462} {(re)construing
  meaning in {NLP}}.
\newblock In \emph{Proceedings of the 58th Annual Meeting of the Association
  for Computational Linguistics}, pages 5170--5184, Online. Association for
  Computational Linguistics.

\bibitem[{Vabalas et~al.(2019)Vabalas, Gowen, Poliakoff, and
  Casson}]{vabalas2019machine}
Andrius Vabalas, Emma Gowen, Ellen Poliakoff, and Alexander~J Casson. 2019.
\newblock \href {https://doi.org/10.1371/journal.pone.0224365} {Machine
  learning algorithm validation with a limited sample size}.
\newblock \emph{PloS one}, 14(11):e0224365.

\bibitem[{Veli{\v{c}}kovi{\'c} et~al.(2018)Veli{\v{c}}kovi{\'c}, Cucurull,
  Casanova, Romero, Li{\`o}, and Bengio}]{velivckovic2018graph}
Petar Veli{\v{c}}kovi{\'c}, Guillem Cucurull, Arantxa Casanova, Adriana Romero,
  Pietro Li{\`o}, and Yoshua Bengio. 2018.
\newblock \href {https://openreview.net/forum?id=rJXMpikCZ} {{G}raph
  {A}ttention {N}etworks}.
\newblock In \emph{International Conference on Learning Representations}.

\bibitem[{Verhagen et~al.(2007)}]{verhagen2007construal}
Arie Verhagen et~al. 2007.
\newblock \href {https://doi.org/10.1093/oxfordhb/9780199738632.013.0003}
  {Construal and perspectivization}.
\newblock \emph{The Oxford handbook of cognitive linguistics}, 48:81.

\bibitem[{Wang et~al.(2019)Wang, Pan, Hu, Long, Jiang, and
  Zhang}]{ijcai2019graphclustering}
Chun Wang, Shirui Pan, Ruiqi Hu, Guodong Long, Jing Jiang, and Chengqi Zhang.
  2019.
\newblock \href {https://doi.org/10.24963/ijcai.2019/509} {Attributed graph
  clustering: A deep attentional embedding approach}.
\newblock In \emph{Proceedings of the Twenty-Eighth International Joint
  Conference on Artificial Intelligence, {IJCAI-19}}, pages 3670--3676.
  International Joint Conferences on Artificial Intelligence Organization.

\bibitem[{Zhou et~al.(2020)Zhou, Dong, Lee, Hooi, Xu, and
  Feng}]{zhou2020nodenorm}
Kuangqi Zhou, Yanfei Dong, Wee~Sun Lee, Bryan Hooi, Huan Xu, and Jiashi Feng.
  2020.
\newblock \href {https://arxiv.org/abs/2006.07107} {Effective training
  strategies for deep graph neural networks}.
\newblock \emph{arXiv preprint arXiv:2006.07107}.

\end{thebibliography}
\bibliographystyle{acl_natbib}

% \appendix

\end{document}